\documentclass[runningheads]{llncs}
\usepackage[T1]{fontenc}
\usepackage{graphicx}
\usepackage{booktabs}
\usepackage{amssymb}
\usepackage{tablefootnote}

\usepackage[misc]{ifsym}

\usepackage{amsmath}
\usepackage{subcaption}
\usepackage{booktabs}
\usepackage{multirow}
\usepackage[dvipsnames]{xcolor}

\definecolor{cor-very-weak}{HTML}{BBBBBB}
\definecolor{cor-weak}{HTML}{EEBD84}
\definecolor{cor-moderate}{HTML}{F47461}
\definecolor{cor-strong}{HTML}{CB2F44}
\definecolor{cor-strong}{HTML}{F47461}
\definecolor{cor-very-strong}{HTML}{8B0000}

\DeclareMathOperator*{\argmin}{arg\,min}
\usepackage{algorithm}
\usepackage{algorithmicx}
\usepackage[noend]{algpseudocode}
\algnewcommand{\LineComment}[1]{\Statex \(\triangleright\) #1}
\algnewcommand\algorithmicinput{\textbf{Input:}}
\algnewcommand\algorithmicoutput{\textbf{Output:}}
\algnewcommand\Input{\item[\algorithmicinput]}
\algnewcommand\Output{\item[\algorithmicoutput]}

\def\algbackskip{\hskip-\ALG@thistlm}

\begin{document}

\title{Fair Clustering with Clusterlets}

\titlerunning{Fair Clustering with Clusterlets}

\author{Mattia Setzu, Riccardo Guidotti}

\institute{University of Pisa, CNR}


\authorrunning{Setzu and Guidotti}


\maketitle              

\begin{abstract}
  Given their widespread usage in the real world, the fairness of clustering methods has become of major interest.
  Theoretical results on fair clustering show that fairness enjoys transitivity: given a set of small and fair clusters, a trivial centroid-based clustering algorithm yields a fair clustering.
  Unfortunately, discovering a suitable starting clustering can be computationally expensive, rather complex or arbitrary.
  
  In this paper, we propose a set of simple \emph{clusterlet}-based fuzzy clustering algorithms that match single-class clusters, optimizing fair clustering.
  Matching leverages clusterlet distance, optimizing for classic clustering objectives, while also regularizing for fairness.
  Empirical results show that simple matching strategies are able to achieve high fairness, and that appropriate parameter tuning allows to achieve high cohesion and low overlap.

\end{abstract}

\section{Introduction}
\label{sec:introduction}
The rapid progress in machine learning has transformed a wide range of fields, enabling algorithms to process intricate data and influencing critical decision-making processes as well as everyday activities~\cite{tufail2023advancements}. 
However, this technological advancement raises concerns about fairness, especially when data-driven grouping unintentionally reinforces pre-existing biases~\cite{DBLP:propublica/AngwinLMK16,DBLP:journals/sciadv/DresselF18}. 
Clustering algorithms, when applied to biased data, can yield unfair partitions that disproportionately affect specific groups and intensify social inequalities. 
This concern is particularly pressing in domains like resource allocation, recruitment, recommendation systems, and medical research, where biased groupings can lead to severe and far-reaching outcomes~\cite{DBLP:journals/access/ChhabraMM21}.

While fairness concerns in machine learning have traditionally focused on supervised learning~\cite{mehrabi2021survey}, recent research has turned attention to unsupervised methods like clustering algorithms~\cite{DBLP:journals/access/ChhabraMM21}. 
In clustering, the objective is to create groups or profiles of instances, maximizing some notion of cluster cohesion and separation.
This task is central to data mining, and has many applications in consequential domains, such as fraud detection~\cite{behera2015credit}, credit scoring~\cite{xiao2016ensemble}, customer profiling~\cite{syakur2018integration}, hiring~\cite{feser2008clusters}, targeted advertisment~\cite{bentolila2013system}, and high-stakes applications such as predictive medicine~\cite{alashwal2019application}.
Similar to biased classification models, unfair clustering practices can reflect, perpetuate, and amplify societal biases, leading to systematic disparate treatment and disproportionately benefiting certain demographic groups over others~\cite{DBLP:conf/nips/BeraCFN19}.

Efforts in tackling this problem propose algorithmic~\cite{DBLP:journals/corr/abs-1802-05733,DBLP:journals/corr/abs-1812-10854,DBLP:conf/nips/HuangJV19,DBLP:journals/jcss/BandyapadhyayFS24}, learned~\cite{DBLP:conf/aaai/ZikoYGA21}, or data augmentation~\cite{DBLP:conf/afci/ChhabraSM21} solutions.
These balance the cohesion of the resulting clustering and its fairness, sometimes providing theoretical guarantees over them~\cite{DBLP:journals/access/ChhabraMM21}.

In this paper, we propose a novel fair clustering framework called \emph{Clusterlets}, which leverages the idea of small, highly cohesive sub-clusters that can be matched in a fairness-aware manner. 
Unlike traditional fair clustering methods that rely on pre-processing or heuristic-based fairness adjustments, our approach introduces a simple and structured matching process that explicitly balances cohesion and deviation.

Our key contributions are as follows.
We introduce \textbf{Clusterlets}, a novel approach to fair clustering that initially generates highly cohesive monochrome clusters and subsequently matches them to minimize deviation.
We define a \textbf{matching-based clustering strategy}, which uses distance and deviation metrics to optimize cluster cohesion while ensuring fairness.
We propose multiple \textbf{pinball-based matching strategies}, including greedy and exhaustive approaches, to improve clustering quality under fairness constraints.
We evaluate our approach on real-world datasets, demonstrating that Clusterlets achieve superior fairness while maintaining strong clustering performance.

The rest of the paper is organized as follows.
In Section~\ref{sec:related} we review related work and presents the necessary concepts to understand our proposal. 
Section~\ref{sec:clusterlets} describes the details of Clusterlets enhancing the difference matching strategies proposed. 
Section~\ref{sec:experiments} presents the experimental results, and Section~\ref{sec:conclusion} concludes with a discussion of future research directions.

\section{Related Work and Background}
\label{sec:related}
In clustering, fairness assumes a definite connotation: a clustering is deemed fair if each cluster is considered fair according to some notion of balance~\cite{DBLP:journals/corr/abs-1802-05733}.
Unlike in supervised learning, where models can directly be regularized to be \textit{locally} fair, in clustering fairness involves sets of instances.
To add to the complexity, a competing objective is at play: clustering cohesion.

Given the plethora of variables at play, and their interconnection, fairness is typically addressed at one of three different times in the clustering pipeline~\cite{DBLP:journals/access/ChhabraMM21}:
\textit{pre}-processing, \textit{in}-processing, and \textit{post}-processing.
At pre-processing time, the data is augmented or manipulated to suit the clustering algorithm: rather than tackling unfairness direclty in the algorithm, one augments the data to the clustering algorithm so that the resulting clustering, performed with classic clustering algorithms, will be a fair one.
In-process approaches, which constitute the bulk of fair clustering solutions, instead engineer the clustering algorithm itself so that the resulting clustering is fair regardless of the data pre-processing.
Finally, post-processing approaches adjust the output of a clustering algorithm so that the resulting clustering is fair.

The variety of processing times is matched by a variety of approaches.
Algorithms like Fairlets~\cite{DBLP:journals/corr/abs-1802-05733} and Coresets~\cite{DBLP:journals/corr/abs-1812-10854,DBLP:conf/nips/HuangJV19,DBLP:journals/jcss/BandyapadhyayFS24}, aim to preprocess the data to reduce fair clustering to standard clustering by computing a set of small and fair clusters, which are then clustered.
Others like antidotes~\cite{DBLP:conf/afci/ChhabraSM21}, instead opt to generate synthetic data, balance the dataset, and then cluster.
More recent approaches leverage neural networks~\cite{DBLP:conf/aaai/ZikoYGA21}, regularizing them for fairness.
Simpler models, such as Decision Trees, are also employed~\cite{DBLP:journals/corr/abs-1910-05113,barata2021fair}.
In the following, we delve in detail into these approaches after recalling basic notions of fairness.\smallskip

At the core of fair clustering is the notion of fairness.
While there exists a plethora of definitions~\cite{DBLP:journals/csur/MehrabiMSLG21}, in fair clustering most works rely on the notion of binary \emph{balance}, which defines the relative frequencies of different groups within a given set.
Informally, given a union $X$ of two non-overlapping and non-empty sets $X_{k_1}, X_{k_2}$, balance defines the relative frequency of one sets with respect to the other.
The two sets may indicate any two groups of interest, e.g., men and women, or discrete values of a sensitive attribute, e.g., political affiliation.
Groups are typically identified by a \textit{color}, and the set of all colors is indicated with $K$.
For illustration purposes, we focus on the bichromatic case where $K = \{ k_1, k_2 \}$.
Formally, we define balance as:
\begin{definition}[Balance]
Given a set $X = A \cup B$, the \textbf{balance} of $X$ is
\begin{equation*}
  \mathit{balance}(X) = \min
  \left(
    \frac{| X_{k_1} |}{| X_{k_2} |},
    \frac{| X_{k_2} |}{| X_{k_1} |}
  \right).
\end{equation*}
\end{definition}
For perfectly balanced sets with an equal color frequency ($| X_{k_1} | = | X_{k_2} |$) we have a unitary balance, which decays to $0$ for perfectly unbalanced sets ($| X_{k_1} | = 1 \bigvee | X_{k_2} | = 1$).
Clustering defines a set of sets, hence we overload the notion of balance to sets of sets:
\begin{equation*}
    \mathit{balance}(X_1, \dots, X_k) = \min \{balance(X_k)\}_{k \in K}.
\end{equation*}
Complementary to the notion of balance is \emph{deviation}:
\begin{definition}[Deviation]
Given two sets $X^1 = \{ X^1_{k_1}, X^1_{k_2} \}, X^2 = \{ X^2_{k_1}, X^2_{k_2} \}$, the \textbf{deviation} $\delta(X^1, X^2)$ of $X^1, X^2$ is defined as
\begin{equation*}
  \delta(X_1, X_2) =
  \frac{| \text{ } | X^1_{k_1} | - | X^2_{k_1} | \text{ } |}{\max | X^1_{k_1} |, | X^2_{k_1} |}.
\end{equation*}
\end{definition}
Deviation measures the difference in relative color frequency, and empyrically the divergence in color distribution.
Generalizing to $n$ colors, deviation is also defined as a metric between color distributions, e.g., Kullback-Leiber divergence.
Balance and deviation are core notions at the basis of most fair clustering algorithms, which we detail below.

\paragraph{Fairlets and Coresets.} Fairlet-based algorithms~\cite{DBLP:journals/corr/abs-1802-05733} apply a two-step clustering algorithm minimizing deviation between the original dataset and each cluster.
The first step is a \textit{fairlet} decomposition: it decomposes the initial dataset $X$ in a set of small clusters, named \textit{fairlets}, respecting the overall dataset balance.
It follows that deviation of each cluster is also minimized.
The latter step is a simple clustering algorithm, either center- or median-based, e.g., $k$-Means~\cite{DBLP:journals/tit/Lloyd82}, which is guaranteed to preserve minimum deviation up to a cost.

The initial fairlet decomposition relies on the construction of a flow graph connecting instances of the two colors.
Edges in the graph are given a weight according to their distance, either $1$, for distances below a user-defined threshold $\tau$, or $+\infty$.
Computing the minimum flow of such graph identifies a set of edges whose weight defines the decomposition cost, and a set of vertice pairs of either color.
The connected instances thus identify the initial fairlets to be clustered.
In fairlet approaches, graph construction relies on an inductive bias, as the subgraphs of vertices of different colors are not strongly connected (cfr.~\cite{DBLP:journals/corr/abs-1802-05733}, Sec. 4.2), thus fairlets are partially predefined.

Coresets~\cite{DBLP:journals/corr/abs-1812-10854,DBLP:conf/nips/HuangJV19,DBLP:journals/jcss/BandyapadhyayFS24} build on top of fairlets by improving on graph construction, which instead considers the complete graph (cfr.~\cite{DBLP:journals/corr/abs-1812-10854}, Algorithm 1).
In Coresets algorithms, edge weights are simply defined by the pairwise distances of the instances defining the edge.
Fairlets and Coresets have seen some improvements on the edges, improving their computational complexity.
Graph construction has been tackled with $\gamma$-trees~\cite{DBLP:conf/icml/BackursIOSVW19}.
Other works instead aim to improve optimality:~\cite{DBLP:conf/nips/BeraCFN19} constructs a linear relaxation of the problem, and starting from an approximate solution, induces an optimal fair clustering.
This comes at a cost in problem size.

\paragraph{Antidotes.} Antidote fair clustering~\cite{DBLP:conf/afci/ChhabraSM21} does away with fairlet or coreset computation, opting for a pre-processing approach.
Antitode augments the original dataset with a set of points, and only then performs clustering.
Points are generated to balance the dataset, so that a downstream clustering algorithm produces, by design, fair clusters.

\paragraph{Clustering Objectives.} A family of in-processing approaches instead relies on altering the clustering objective of standard clustering algorithms, integrating deviation minimization.
Fair Round-Robin Algorithm for Clustering (FRAC)~\cite{DBLP:journals/datamine/GuptaGKJ23}, enforces deviation minimization fairness by solving a fair assignment problem during each iteration of $k$-means clustering. 
Similar approaches are implemented in FairKM~\cite{DBLP:conf/edbt/Abraham0S20} and Fair-Lloyd~\cite{DBLP:conf/fat/GhadiriSV21}.

\paragraph{Neural Clustering.} Neural networks find successful applications also in clustering.
In variational fair clustering~\cite{DBLP:conf/aaai/ZikoYGA21}, the clustering and fairness objective define the training loss of a neural model.
While such model is in theory  more powerful, as the two objectives can be regulated to the desired degree, it lacks the strong theoretical backing of both Fairlet and Coreset approaches.
Computational time and high variance of the algorithm are also primary concerns.

\paragraph{Tree Fair Clustering.} Other in-process fair clustering algorithms rely on leveraging Decision Trees to partition, and thus cluster, the space.
These algorithms include the Fair Tree Classifier and the Splitting Criterion AUC for Fairness (SCAFF), introduced in~\cite{DBLP:journals/corr/abs-1910-05113} and~\cite{barata2021fair}, respectively. 
The former introduces a novel combination of threshold-free demographic parity with ROC-AUC optimization, which supports multiple, multicategorical, or intersectional protected attributes.

\bigskip
No single fairness measure is universally optimal for clustering, which underscores the importance of adopting a balanced approach tailored to the dataset and clustering objectives. 
Furthermore, the formalization of these fairness measures varies significantly across different studies.

Our proposal, Clusterlets, borrows the core idea of cluster decomposition from Fairlets and Coresets.
Unlike both of them, it leverages a matching algorithm to aggregate them into clusters.
Like neural clustering, its hyperparameters can regulate the cost of fairness, thus providing flexibility.
Like FRAC, it provides an in-processing approach to fair clustering.

\begin{algorithm}[t]
    \begin{algorithmic}[1]
    \Input $| K |$ monochrome datasets $\{ X^1, \dots, X^{| K |} \}$, clustering hyperparameters $\theta_C$, matching hyperparameters $\theta_M$
    \Output A clustering $C$
    \State $C \gets \bigcup\limits_{X \in X^K} f_C(X, \theta_C)$
    \State $C \gets $ \Call{Match}{$C$, $\theta_M$}
    \State \Return{$C$}
    \end{algorithmic}
    \caption{Clusterlets algorithm: first, monochrome clusterlets are extracted from each color (Line 1); then, they are matched (Line 2).}
    \label{alg:clusterlets}
\end{algorithm}

\section{Clusterlets}
\label{sec:clusterlets}
In this section, we present Clusterlets, our proposal for fairness clustering.
Similarly to Fairlets~\cite{DBLP:journals/corr/abs-1802-05733}, Clusterlets first generates a set of initial small clusters, which we call \emph{clusterlets}, then applies an aggregation to generate the output clustering.
Both Fairlets and Clusterlets implement the former with pairwise distances among instances, constructing graphs which, by design of their edges, minimize deviation.
This process, while akin to clustering, constraints the initial solution to minimize deviation at a potential cost in clustering cohesion.
Instead, in our approach we reverse this process, tackling clustering cohesion first, and deviation second.
The core idea behind this is rather simple: look for highly cohesive clusters first, and only then aggregate them to minimize deviation.
Thus, we push the search for initial cohesive cluster approach to its natural extreme, and directly rely on a foundational, simpler, and well-studied approach for highly cohesive cluster discovery, i.e., clustering itself.
Unlike both Fairlets and Coresets, Clusterlets relies on \textit{matching} to aggregate the initial set of clusters, instead of on clustering.
A matching strategy looks to grow clusters by iteratively finding suitable clusterlets.
This approach allows us to have more flexibility, and optimize for deviation while maintaining cohesive clusters.


\smallskip
We detail our proposal, Clusterlets, in Algorithm~\ref{alg:clusterlets}, following a straightforward two-step process:
\begin{itemize}
    \item \textbf{Clusterlet Extraction.} For each color, it \textit{extracts} a set of \textit{monochrome} clusters $C = \{c_1,$ $ \dots,$ $c_N\}$, named clusterlets (line 1)
    \item \textbf{Clusterlet Matching.} Such monochrome clusters are \textit{matched}, yielding the final clustering $C$ (line 2)
\end{itemize}
Once again, with the word ``color'' we refer to a category, i.e., to a value of a categorical attribute, e.g. ``male'' or ``female'' for ``sex'', etc.
Thus, with monochrome cluster we refer to a set of instances belonging to the same category.

The extraction of monochrome clusters yields sets of monochrome \emph{clusterlets} $\{ c^{k^i} \}_{k^i \in K}$.
We use $c^{k^i}$ to indicate the monochrome set of clusterlets of color $k^i$, and $c^{k^i}_j$ to indicate a clusterlet in this set.
Given that such clusterlets are extracted from a clustering algorithm, they already retain high cohesion but being monochrome, they also have a high deviation.
This process is defined by a clustering algorithm $f_C$, e.g., $k$-means, parametrized by a parameters $\theta_C$ (Line 1), e.g., the number of clusters $k$.

To then construct polychrome clusters, we rely on \emph{matching}, to finally construct balanced clusters out of this set of monochrome clusterlets (Line 2 of Algorithm~\ref{alg:clusterlets}).
Matching aims, for a given clusterlet $c$, to find an appropriate (set of) clusterlet $c'$ minimizing the deviation and cohesion of $c \cup c'$.
Notably, matching itself is parametric, and different matching strategies can be defined to suit different needs, either favoring high cluster cohesion, or lower deviation.
Moreover, matching allows us to generate \textit{fuzzy} clusters, as a single clusterlet can be matched by several other clusterlets.
While crisp clustering is more interpretable, in an unbalanced setting in which fair clustering is required, crispness poses an additional constraint which might negatively impact clustering cohesion.

\subsection{Clusterlet Extraction}
Color-specific clusterlet extraction leverages $k$-means clustering, which extracts $k$ clusters out of the given monochrome dataset.
Here, we can vary $k$ to achieve finer-grained clusterlets, which ease matching in the following step of the algorithm.
This comes at a computational cost, introducing a trade-off between the computational cost of the clustering algorithm and the performance of the downstream matching.

\begin{algorithm}[t]
    \begin{algorithmic}[1]
    \Input Maximum number of hops $hops$
    \Output A clustering $C$
    \State $\hat{C} \gets \{ \}$
    \For{$c \in c^{k^1}$}  \Comment{Clusterlet of the first color}
    \State $\hat{C} \gets \{ c \}$
    \For{$hop \in \{ 1, \dots,  hops\}$} \Comment{Hopping over all colors}
      \If{$hop \mod 2 = 1$}
        \For{$i \in \{ 2, \dots, | K | \} $} \Comment{Forward hop}
          \State $\hat{C} \gets \hat{C} \cup \{ \argmin\limits_{c^{k^i}} f_M(c', \theta_M) \}$ 
        \EndFor
        \Else
        \For{$i \in \{ | K |, \dots, 2 \} $} \Comment{Backward hop}
          \State $\hat{C} \gets \hat{C} \cup \{ \argmin\limits_{c^{k^i}} f_M(c', \theta_M) \}$ 
        \EndFor
      \EndIf
      \EndFor
      \State $\hat{C} \gets C \cup \hat{C}$ 
    \EndFor
    \State \Return{$C$}
    \end{algorithmic}
    \caption{
        Matching monochrome clusterlets: starting from a color $k^1$, we add clusterlets of different colors (Lines 6 and 10), until a maximum number of iterations ($hop$s) is reached (Line 4).
    Each hop is performed from the latest clusterlet, thus to \textit{forward} hops (from $k^1$ to $k^{| K |}$, Line 6) follow \textit{backward} hops (from $k^{| K |}$ to $k^1$, Line 9).
    The process is repeated for every clusterlet of $c^{k^1}$ (Line 2).
    }
    \label{alg:matching}
\end{algorithm}

\subsection{Clusterlet Matching}
Matching monochrome clusterlets involves optimizing both cohesion and deviation.
Informally, a match joins an existing cluster with a monochrome clusterlet, creating a new cluster.
While the cohesion is already locally minimized in each monochrome clusterlet, growing the cluster is bound to
bring it back up.
The opposite is true for deviation, which is maximal in each monochrome clusterlet.
The two metrics may pull in different directions, and matching one clusterlet over the other may minimize loss in cohesion while not minimizing the reduction in deviation.
Rather than arbitrarily defining an optimal match, we propose a set of matching algorithms, each tackling a different approach.

Inspired by Fairlets and Coresets, we create a $k$-partite graph, each monochro\-me clusterlet associated to a node, and each partition associated to a color.
Edges define the distance between centroids of each clusterlets.
The matching algorithms leverage this graph representation by looking for a clusterlet minimizing some loss in other partitions.
This process is repeated back and forth across the partitions, as matching algorithms bounce back and forth across different colors.
Thus, we name these \textit{pinball} matchers.

\subsubsection{Pinball Matchers.}
In a pinball matcher, monochrome clusterlets are first divided into $K$ sets according to their color.
Then, as detailed in Algorithm~\ref{alg:matching}, starting from a clusterlet $c$ of a given color (Line 1), we find a matching clusterlet $c'$ (Lines 6 and 10) of the next color optimizing some measure, e.g., cohesion.
$c'$ is then added to the current cluster.
This process is repeated until all colors are exhausted (Lines 6 and 9), resulting in one \emph{hop} across the color set.
Each hop iteratively grows the cluster in the opposite color direction until a maximum number of hops is reached (Line 4), and a cluster is defined (Line 11).

Note that hopping between partitions may end up selecting the same monoch\-rome clusterlet more than once, hence two clusters may well include the same single monochrome clusterlet.
Thus, unlike Fairlet and Coreset clustering, matching may create fuzzy clusters, i.e., instances may be assigned to multiple clusters.
While this may create unseparable clusters, it allows us to increase cluster cohesion, and to lower deviation in highly imbalanced and dense datasets.

Finally, the last component to consider for a pinball matcher is the \textit{matching measure}, of which we use two:
\begin{itemize}
\item \textit{Distance.} Among a set of candidate clusterlets, the closest one is chosen. 
\item \textit{Deviation.} Among a set of candidate clusterlets, the one minimizing deviation is chosen. Here, we consider the deviation of the constructed cluster, i.e., for a partial cluster $\hat{C}$, we evaluate $\delta(\hat{C} \cup \{ c' \}, X)$ of all monochrome clusterlets $c'$ in a set of candidates.
\end{itemize}

By its own nature, a pinball approach lends itself to both greedy and exhaustive approaches.
In the former, the clusterlet minimizing distance to/deviation w.r.t. the latest added clusterlet is considered; in the latter, the measure is considered over the whole current cluster.

\smallskip
Finally, we can now define a set of matchers:
\begin{itemize}
  \item \textbf{Distance Pinball Matcher} (D): Adds monochrome clusterlet with minimum \textit{cumulative} distance from the clusterlet in the current cluster
  \item \textbf{Greedy Distance Pinball Matcher} (G-D): Adds monochrome clusterlet with minimum distance from \textit{the last added monochrome clusterlet}
  \item \textbf{Balance Pinball Matcher} (B): Adds monochrome clusterlet with minimum deviation from the current cluster
  \item \textbf{Greedy Balance Pinball Matcher} (G-B): Adds monochrome clusterlet minimizing deviation with respect \textit{to the last added monochrome clusterlet}
\end{itemize}

\subsubsection{Centroid Matcher.}
Fair clustering is an NP-hard problem, thus enumerating all possible solution is unfeasible.
This holds true even when considering simple matching, as the solution space is defined as the set of all possible monochrome clusterlet partitions\footnote{The set of all partitions, of any size, of a given set of cardinality $n$, has cardinality given by the $n^{th}$ Bell number. This series is factorial in nature, thus untractable even for a low $n$.}.
We also propose to sample such exponentially large space, evaluating clusterings according to a simple affine combination of the clustering silhouette $s(C)$, and the maximum deviation $\max \delta(c', X)$ of any of its clusterlets:
\begin{equation}
\omega s(C) + (1 - \omega \max\limits_{c' \in C} \delta(c', X)).
\label{eq:centroidmatcher}
\end{equation}
where the $\omega$ hyper-parameter is controlling the level of cohesion.
Indeed, note that here, unlike in previous matchers, we are defining crisp clusters.
Fuzzy clustering would only enlarge the already untractably large set of possible solutions, thus increasing the variance of the sampled solutions.
We name this matcher \textbf{Centroid Matcher} (C).

\section{Experiments}
\label{sec:experiments}
We analyze Clusterlets on three levels: first, the overall algorithm, and its overall performances; second, its sensitivity to hyperparameters; and third, its performance with respect to competitors.
Specifically, we try to answer these questions:
\begin{itemize}
  \item \textbf{Q1.} Is there a best matcher?
  \item \textbf{Q2.} Does matching improve on classic fair clustering algorithms?
  \item \textbf{Q3.} Can we relax fairness in favor of cohesion?
\end{itemize}
In the following, we first present the experimental settings, and then a set of experiments answering these question.
  
\begin{table}[t]
\centering
\setlength{\tabcolsep}{1.2mm}
\begin{tabular}{@{} c c l c l r c c c c @{}}
  \toprule
  & \multicolumn{3}{c}{\textbf{Deviation}  $\delta$ $\downarrow$} & & && \\
  \cmidrule{2-4}
  \textbf{Matcher} & $\mu \pm \sigma$ && $\max$ && \textbf{Size} && \textbf{Cohesion} $\uparrow$ && \textbf{Overlap} $\downarrow$ \\
  \midrule
  C       & $.029 \pm .055$ && $.049 \pm .079$ && $2.04\pm 0.2$    && $.836 \pm .113$ && $.000 \pm .000$ \\
  G-B     & $.053 \pm .050$ && $.204 \pm .193$ && $11.6 \pm 9.0$   && $.864 \pm .100$ && $.000 \pm .000$ \\
  D-PB    & $.113 \pm .085$ && $.264 \pm .165$ && $17.5 \pm 14.2$  && $.886 \pm .091$ && $.295 \pm .243$ \\
  B-PB    & $.114 \pm .060$ && $.385 \pm .244$ && $18.6 \pm 15.2$  && $.862 \pm .102$ && $.287 \pm .218$ \\
  G-D-PB  & $.246 \pm .097$ && $.434 \pm .191$ && $11.2 \pm 8.6$   && $.869 \pm .093$ && $.000 \pm .000$ \\
  \bottomrule
  \end{tabular}
  \caption{Clusterlet results, by matcher. Results averaged over $10$ random seeds, datasets, and hyperparameter configuration.}
  \label{tbl:matcher_results}
\end{table}

\begin{table}[t]
\centering
\setlength{\tabcolsep}{1.3mm}
\begin{tabular}{@{} c c r r c c l c r @{}}
  \toprule
  & & & & \multicolumn{2}{c}{\textbf{Deviation}  $\downarrow$} && \\
  \cmidrule{5-6}
  \textbf{Dataset} & \textbf{Matcher} & $k$ & \textbf{Size} & $\mu \pm \sigma$ & $\max$ && \textbf{Cohesion} $\uparrow$ & \textbf{Overlap} $\downarrow$\\
    \midrule
    adult 	            & G-B &	$20$ 	& $11$ 	& $.004 \pm .004$ & $.015$ && $.923$ & $0.0$\\
    credit 	            & G-B &	$5$ 	& $3$ 	& $.008 \pm .006$ & $.015$ && $.893$ & $0.0$\\
    bank 	              & G-B &	$30$ 	& $17$ 	& $.000 \pm .001$ & $.004$ && $.978$ & $0.0$\\
    breast 	            & G-B &	$5$ 	& $3$ 	& $.002 \pm .000$ & $.002$ && $.551$ & $0.0$\\
    compas 	            & G-B &	$5$ 	& $3$ 	& $.004 \pm .004$ & $.010$ && $.936$ & $0.0$\\
    heart failure 	    & G-B &	$5$ 	& $4$ 	& $.003 \pm .002$ & $.006$ && $.943$ & $0.0$\\
    heloc 	            & G-B &	$20$ 	& $17$ 	& $.008 \pm .007$ & $.030$ && $.809$ & $0.0$\\
    magic 	            & G-B &	$5$ 	& $3$ 	& $.005 \pm .003$ & $.008$ && $.868$ & $0.0$\\
    nbfi 	              & G-B &	$20$ 	& $13$ 	& $.001 \pm .001$ & $.004$ && $.922$ & $0.0$\\
    pima 	              & G-B &	$5$ 	& $3$ 	& $.003 \pm .001$ & $.005$ && $.880$ & $0.0$\\
    speeddating 	      & G-B &	$5$ 	& $4$ 	& $.004 \pm .003$ & $.009$ && $.710$ & $0.0$\\
  \bottomrule
  \end{tabular}
  \caption{Best matcher (according to mean deviation) per dataset. Limit cases (only $2$ clusters, or overlap degree $\geq 0.5$) filtered out.}
  \label{tbl:matcher_selection_mean_deviation}
\end{table}

\begin{table}[t]
\centering
\setlength{\tabcolsep}{1.3mm}
\begin{tabular}{@{} l c r r l c c c c c c c @{}}
  \toprule
  & & & && \multicolumn{2}{c}{\textbf{Deviation}  $\delta$ $\downarrow$} & & &&& \\
  \cmidrule{6-7}
  \textbf{Dataset} & \textbf{Matcher} & $k$ & \textbf{Size} && $\mu \pm \sigma$ & $\max$ && \textbf{Cohesion} $\uparrow$ && \textbf{Overlap} $\downarrow$ \\
  \midrule
    adult 	            & G-D-PB 	& $5$ 	& $4$ 	&& $.239 \pm .000$ &	$.239$ &&	$.923$ && $.000$\\
    credit 	& D-PB 	  & $5$ 	& $9$ 	&& $.388 \pm .134$ &	$.468$ &&	$.988$ && $.449$\\
    bank 	            & G-B 	  & $5$ 	& $4$ 	&& $.037 \pm .047$ &	$.119$ &&	$.982$ && $.000$\\
    breast 	            & D-PB 	  & $30$ 	& $53$ 	&& $.285 \pm .110$ &	$.437$ &&	$.791$ && $.100$\\
    compas 	            & G-D-PB 	& $10$ 	& $7$ 	&& $.222 \pm .179$ &	$.640$ &&	$.961$ && $.000$\\
    heart failure 	    & D-PB 	  & $30$ 	& $38$ 	&& $.116 \pm .084$ &	$.345$ &&	$.988$ && $.290$\\
    heloc 	            & B-PB 	  & $20$ 	& $30$ 	&& $.099 \pm .091$ &	$.322$ &&	$.811$ && $.297$\\
    magic 	            & D-PB 	  & $30$ 	& $40$ 	&& $.116 \pm .088$ &	$.281$ &&	$.928$ && $.403$\\
    nbfi 	            & D-PB 	  & $30$ 	& $35$ 	&& $.024 \pm .015$ &	$.056$ &&	$.953$ && $.306$\\
    pima 	            & D-PB 	  & $30$ 	& $40$ 	&& $.123 \pm .110$ &	$.401$ &&	$.958$ && $.214$\\
    speeddating 	    & D-PB 	  & $30$ 	& $41$ 	&& $.101 \pm .086$ &	$.393$ &&	$.793$ && $.436$\\
  \bottomrule
  \end{tabular}
  \caption{Best matcher (according to cohesion) per dataset. Limit cases (only $2$ clusters, or overlap degree $\geq 0.5$) filtered out.}
  \label{tbl:matcher_selection_cohesion}
\end{table}

\begin{table}[t!]
\centering
\begin{tabular}{@{} c c c c c c c c c c @{}}
  \toprule
  && & & & \multicolumn{2}{c}{\textbf{Deviation}  $\downarrow$} & \\
  \cmidrule{6-7}
  \textbf{Dataset} & \textbf{Method} & \textbf{Matcher} & $k$ & \textbf{Size} & $\mu \pm \sigma$ & $\max$ && \textbf{Cohesion} $\uparrow$ & \textbf{Overlap} $\downarrow$\\
    \midrule
    \multirow{2}{*}{adult} 	            & Clusterlet & G-B &	$20$ 	& $11$ 	& $.004 \pm .004$ & $.015$ && $.923$ & $0.0$\\
    & FRAC & & & $5$ & $.025 \pm .023$ & $.063$ && $.680$ & $0.0$ \\
    \midrule
    \multirow{2}{*}{credit} & Clusterlet	& G-B &	$5$ 	& $3$ 	& $.008 \pm .006$ & $.015$ && $.893$ & $0.0$\\
    & FRAC & 	& & $10$ 	& $.205 \pm .156$ & $.480$ && $.986$ & $0.0$ \\
    \midrule
    \multirow{2}{*}{bank} 	              & Clusterlet	& G-B &	$30$ 	& $17$ 	& $.000 \pm .001$ & $.004$ && $.978$ & $0.0$\\
    & FRAC & 	& & $10$ 	& $.020 \pm .015$ & $.0518$ && $.979$ & $0.0$ \\
    \midrule
    \multirow{2}{*}{breast} 	            & Clusterlet	& G-B &	$5$ 	& $3$ 	& $.002 \pm .000$ & $.002$ && $.551$ & $0.0$\\
    & FRAC & 	& & $5$ 	& $.215 \pm .134$ & $.3966$ && $.635$ & $0.0$ \\
    \midrule
    \multirow{2}{*}{compas} 	            & Clusterlet	& G-B &	$5$ 	& $3$ 	& $.004 \pm .004$ & $.010$ && $.936$ & $0.0$\\
    & FRAC & 	& & $5$ 	& $.080 \pm .047$ & $.1481$ && $.963$ & $0.0$ \\
    \midrule
    \multirow{2}{*}{heart failure} 	    & Clusterlet	& G-B &	$5$ 	& $4$ 	& $.003 \pm .002$ & $.006$ && $.943$ & $0.0$\\
    & FRAC & 	& & $5$ 	& $.034 \pm .028$ & $.0751$ && $.956$ & $0.0$ \\
    \midrule
    \multirow{2}{*}{heloc} 	            & Clusterlet	& G-B &	$20$ 	& $17$ 	& $.008 \pm .007$ & $.030$ && $.809$ & $0.0$\\
    & FRAC & 	& & $5$ 	& $.092 \pm .060$ & $.1717$ && $.809$ & $0.0$ \\
    \midrule
    \multirow{2}{*}{magic} 	            & Clusterlet	& G-B &	$5$ 	& $3$ 	& $.005 \pm .003$ & $.008$ && $.868$ & $0.0$\\
    & FRAC & 	& & $5$ 	& $.047 \pm .048$ & $.1265$ && $.910$ & $0.0$ \\
    \midrule
    \multirow{2}{*}{nbfi} 	              & Clusterlet	& G-B &	$20$ 	& $13$ 	& $.001 \pm .001$ & $.004$ && $.922$ & $0.0$\\
    & FRAC & 	& & $5$ 	& $.011 \pm .005$ & $.0190$ && $.933$ & $0.0$ \\
    \midrule
    \multirow{2}{*}{pima} 	              & Clusterlet	& G-B &	$5$ 	& $3$ 	& $.003 \pm .001$ & $.005$ && $.880$ & $0.0$\\
    & FRAC & 	& & $5$ 	& $.135 \pm .086$ & $.2686$ && $.922$ & $0.0$ \\
    \midrule
    \multirow{2}{*}{speeddating} 	      & Clusterlet	& G-B &	$5$ 	& $4$ 	& $.004 \pm .003$ & $.009$ && $.710$ & $0.0$\\
    & FRAC & 	& & $5$ 	& $.019 \pm .015$ & $.0422$ && $.739$ & $0.0$ \\
    \midrule
  \bottomrule
  \end{tabular}
  \caption{Clusterlet (top row of each block) VS FRAC (bottom of each block). Limit cases (only $2$ clusters, or overlap degree $\geq 0.5$) filtered out.}
  \label{tbl:FRAC_VS_Clusterlets}
\end{table}
  
\begin{figure}[t]
    \centering
    \includegraphics[width=0.75\textwidth]{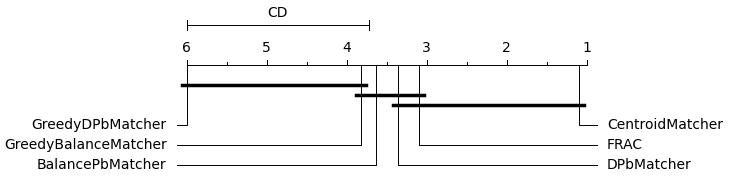}
    \caption{Critical difference plot of different matchers. Each entry in the rank line shows the average rank of a matcher (right is better). Matchers whose differences are not statistically significant are connected by a black band.}
    \label{fig:matcher_cdplot}
\end{figure}
\begin{table}[t]
\centering
\begin{tabular}{@{} r r r r l r r r r l r r r r @{}}
  \toprule
  & \multicolumn{3}{c}{\textbf{Pearson}} && \multicolumn{3}{c}{\textbf{Spearman}} && \multicolumn{3}{c}{\textbf{Kendall}} \\
  \cmidrule{2-4}
  \cmidrule{6-8}
  \cmidrule{10-12}
                & $s$ & $\omega$ & $\delta$ && $s$ & $\omega$ & $\delta$ && $s$ & $\omega$ & $\delta$ \\
  \midrule
  $s$           & 1       & 0.449   & 0.029     && 1 & 0.438 & 0.375  && 1       & 0.330   & 0.246  \\
  $\omega$      &         & 1       & 0.284     &&   & 1     & 0.565  &&         & 1       & 0.453  \\
  $\delta$      &         &         & 1         &&   &       & 1      &&         &         & 1       \\
  \bottomrule
\end{tabular}
\caption{Correlations between hyperparameters (deviation $\delta$, cohesion $s$, and silhouette weight $\omega$) and results.}
\label{tbl:correlation_hyperparameters}
\end{table}

\smallskip
\textbf{Experimental Settings.} Due to the fuzzy nature of the pinball matchers, rather than silhouette, for them we report \emph{relative} \textbf{Cohesion} computed as the ratio between the mean inter-cluster distance, and the maximum dataset distance.
For a clustering, this is averaged across clusters.
We finally flip this measures by subtracting it to $1$, thus mapping it to the interval $[0, 1]$.
High values, which indicate highly compact clusters with respect to the dataset, are better.
We accompany this metric with the \textbf{Overlap} \emph{degree}, i.e., the proportion of instances belonging to multiple clusters.
Unlike classical adaptation of silhouettes to fuzzy clustering, jointly these two measures capture both compactness and overlap of the clustering.
For the datasets analyzed, we use the class as color.

In order to find the best hyperparameter configuration for each dataset, we perform a grid search over the following algorithms' hyper-parameters where PB stands for ``Pinball'': 
\textbf{Matcher}: Centroid (C), DistancePB (D), BalancePB (B), GreedyBalance (G-B), GreedyDistancePB (G-D);
\textbf{Hops}: 1, 2, 3, 4;
\textbf{Cohesion $\omega$}: 0, 0.25, 0.5, 0.75;
\textbf{Number of clusters $k$}: 2, 5, 10, 20, 30; 
\textbf{Sample size}: $2.5 \cdot 10^5$.
Since the clusterlet extraction relies on randomly sampled initial centroids, each grid configuration is repeated $10$ times, each with a different random initialization.
The best configuration for each dataset is reported in Table~\ref{tbl:matcher_selection_mean_deviation}.
We compare Clusterlet only against FRAC~\cite{DBLP:journals/datamine/GuptaGKJ23}, for which we use the same hyperparameter space.
Public implementations of Fairlet and Coresets algorithms were either lacking or not working.
An implementation of Clusterlets is available on Github.\footnote{Anonymized for submission: https://anonymous.4open.science/r/clusterlets-5B52/README.md Datasets, all from the UCI or Kaggle repository, are not included in the software repository simply to avoid overloading the repository.} \smallskip
  
\textbf{[Q1.] Matching Algorithms.} By analyzing Table~\ref{tbl:matcher_results} we observe that matching algorithms show markedly different results, the best performing of which being Centroid Matcher (C), followed by the Greedy Balance Matcher (G-B).
While the latter has a larger mean deviation ($\approx 1,5 \times$), it has also a comparatively much lower variance ($\approx 189\%$ of the mean VS $96\%$), indicating a lower algorithmic variance.
This is to be expected, as the Balance Matcher does not rely on stochastic sampling, unlike the CentroidMatcher.
The large mean deviation difference between the two is statistically significant, as shown in the critical difference plot in Figure~\ref{fig:matcher_cdplot}.
Surprisingly, mean deviation-wise, the CentroidMatcher is instead equivalent to the Distance Pinball Matcher (D), which instead relies on pure distance to match clusterlets.
Still, such matcher is itself equivalent to the other balance-based (and thus deviation-based) matchers.
While, as expected, the Centroid Matcher is an upper bound of all matchers, we do not generally find a great separation among different matchers.
    
\smallskip
\textbf{[Q1.] Model Selection.} Focusing on single models, Table~\ref{tbl:matcher_selection_mean_deviation} reports the best clusterings on each dataset, selected by mean deviation.
Here, we have filtered edge cases with a low number of clusters ($< 3$), and high overlap degree ($> 0.5$).
The first filter in particular yields significantly different results.
On each dataset, the Greedy Balance Matcher (G-B) is the best performing one.
Of interest are two factors: first, the cohesion.
With one notable exception (breast dataset), cohesion is high, suggesting that deviation and cohesion can be jointly optimized.
Second, the overlap degree, which is extremely low.
This suggests that a fuzzy clustering is not necessary to achieve both high cohesion and low deviation.

Table~\ref{tbl:matcher_selection_cohesion} instead shows the best model, per dataset, according to the highest cohesion.
Here we see a shift in matchers, with Pinball matchers being the top performing ones across the board.
This also come in a shift in clustering size, markedly higher than in the previous case, as well as in overlap and deviation.
Strictly optimizing for cohesion appears to worsen the remaining metrics, while optimizing for deviation instead achieves good measures across the board.

\smallskip
\textbf{[Q2.] Cohesion.} Focusing on Centroid Matcher, which offers the hyperparameter $\omega$ to control cohesion, we observe that when selecting for cohesion, as we show in Table~\ref{tbl:matcher_selection_mean_deviation}, we have a significant hyperparameter shift.
As expected, as we increase the weight given to silhouette (hyperparameter $\omega \in [0, 1]$), the silhouette score of the resulting clustering increases.
The trend generally holds, as mean deviation is negatively correlated with $\omega$ and, as a consequence, silhouette as can be seen in Table~\ref{tbl:correlation_hyperparameters}.
Moreover, mean and maximum deviation still retain very low values, with only one pathological case (breast dataset) achieving a mean deviation over $10\%$, while on all other datasets we achieve a deviation lower of at least one order of magnitude.

\smallskip
\textbf{[Q3.] Clusterlets VS FRAC.} 
In Table~\ref{tbl:FRAC_VS_Clusterlets} we compare the performance of Clusterlets with FRAC.
These results, and confirmed through critical difference (Figure~\ref{fig:matcher_cdplot}), show that our matching approach reaches state of the art performance.
Deviation is often consistenly lower, while deviation variance tends to be relatively higher.
In some cases, the difference is of an order of magnitude.
Notably, Clusterlets tends to find larger clusterings, while maintaining a comparable, cohesion.
Overlap, which may impair interpretability, is extremely low for Clusterlets as well, even though it was not directly optimized for.

\section{Conclusions}
\label{sec:conclusion}
In this paper, we introduced \textbf{Clusterlets}, a novel fair clustering framework that leverages structured matching of small monochrome clusters to optimize both fairness and clustering cohesion. Our approach builds on the theoretical insights of fair clustering while introducing a practical and effective mechanism for ensuring fairness.

Our empirical results demonstrate that simple yet effective matching strategies can achieve high fairness while maintaining strong cohesion and overlap scores, striking a balance between fairness constraints and clustering quality.
Moreover, the flexibility of our matching framework allows for adaptation to different fairness notions and application domains.

Future work can explore extensions in several directions: \textit{(1)} improving computational efficiency for large-scale datasets;
\textit{(2)} integrating deep learning-based representations for better initialization of Clusterlets;
and \textit{(3)} generalizing the approach to other fairness-sensitive tasks beyond clustering, such as community detection.
By shifting the focus from post-hoc fairness adjustments to fairness-aware clustering from the ground up, Clusterlets offer a promising pathway toward more equitable machine learning applications.

%
%
%
\bibliographystyle{splncs04}
\bibliography{paper}
%

\end{document}